\documentclass[11pt]{article}

\usepackage[final]{acl}

\usepackage{times}
\usepackage{latexsym}

\usepackage[T1]{fontenc}

\usepackage[utf8]{inputenc}

\usepackage{microtype}

\usepackage{inconsolata}

\usepackage{graphicx}
\usepackage{booktabs}
\usepackage{multirow}

\usepackage{listings}
\usepackage{xcolor}
\usepackage{float}

\usepackage{amsmath}
\usepackage{amssymb}

\lstdefinestyle{promptstyle}{
  basicstyle=\ttfamily\scriptsize,
  breaklines=true,
  breakatwhitespace=true,
  columns=fullflexible,
  frame=single,
  framerule=0.3pt,
  keepspaces=true,
  xleftmargin=2pt,
  xrightmargin=2pt,
  aboveskip=2pt,
  belowskip=2pt
}

\title{PIVOT: Perplexity-Informed KD-to-RL Transition Scheduling for Vertical-Domain Few-Shot Distillation}

\author{
  \textbf{Heng Li}\textsuperscript{1,2,\textdagger},
  \textbf{Yong Zhang}\textsuperscript{1,\textdagger},
  \textbf{Ning Cheng}\textsuperscript{1,\textasteriskcentered},
  \textbf{Zhigen Li}\textsuperscript{1}, \\
  \textbf{Yun Zhu}\textsuperscript{1},
  \textbf{Yanmeng Wang}\textsuperscript{1},
  \textbf{Shaojun Wang}\textsuperscript{1},
  \textbf{Jing Xiao}\textsuperscript{1}, \\
  \textsuperscript{1}Ping An Technology (Shenzhen) Co., Ltd., China \\
  \textsuperscript{2}University of Science and Technology of China \\
  \texttt{\{zhangyong203, chengning211\}@pingan.com.cn}
}

\begin{document}
\maketitle

\setlength{\footnotesep}{0.4\baselineskip}

\renewcommand{\thefootnote}{}\setcounter{footnote}{0}
\footnotetext{\textdagger\ Equal contribution.}
\footnotetext{\textasteriskcentered\ Corresponding author.}\setcounter{footnote}{0}
\footnotetext{This work was done during Heng Li's internship at Ping An Technology.}\setcounter{footnote}{0}

\renewcommand{\thefootnote}{\arabic{footnote}}

\begin{abstract}


Vertical-domain few-shot classification remains challenging for small language models, as limited supervision makes it difficult to acquire domain-specific decision knowledge. 
On-Policy Distillation (OPD) can improve teacher-guided adaptation by supervising student-generated rollouts, while GRPO-based reinforcement learning can further refine downstream predictions. 
However, existing KD-to-RL pipelines typically rely on globally fixed transition schedules, ignoring that different samples may require different amounts of teacher-guided acquisition before reward-driven refinement. We propose PIVOT (\textbf{P}erplexity-\textbf{I}nformed Transition \textbf{O}ptimization), a dynamic transition framework that routes samples between OPD and GRPO according to teacher-evaluated sequence perplexity. 
PIVOT moves low-perplexity samples to GRPO for reward-driven refinement while keeping high-perplexity samples under OPD for continued domain knowledge acquisition. Experiments on Banking77 and HWU64 show that PIVOT consistently outperforms continued OPD and globally synchronized OPD$\rightarrow$GRPO baselines under the same number of post-warm-up student optimization steps, achieving stronger downstream performance and more stable training dynamics.

\end{abstract}


\section{Introduction}

In vertical-domain few-shot classification tasks, small language models often struggle to adapt under limited supervision due to insufficient domain-specific knowledge. This challenge is particularly severe in professional domains such as finance, where downstream prediction depends heavily on specialized terminology and domain semantics. As a result, effective adaptation increasingly relies on transferring knowledge from expert teacher models through distillation-based optimization \cite{hinton2015distilling,sanh2019distilbert}. Among existing approaches, On-Policy Distillation (OPD) has emerged as an effective paradigm for teacher-guided adaptation by aligning supervision with student-generated trajectories during optimization, thereby improving exposure alignment during autoregressive generation \cite{agarwal2024gkd,gu2024minillm}.

Recent studies have further explored how knowledge distillation and reinforcement learning should be coordinated during training. GRPO has also been applied to intent detection with chain-of-thought reasoning and reward-based curriculum sampling, improving generalization beyond supervised fine-tuning \cite{feng2025improvinggeneralizationintentdetection}.
RLKD and KDRL \cite{xu2025rlkd,xu2025kdrl} show that reinforcement learning can improve downstream decision refinement beyond teacher-guided imitation. CoDistill-GRPO \cite{kwon2026codistill} and related hybrid optimization frameworks \cite{li2026unifying,ding2026hdpo} further demonstrate that jointly coordinating distillation and RL objectives can improve optimization efficiency and rollout quality. Meanwhile, recent OPD studies have found that prolonged distillation may introduce unreliable supervision or unstable training dynamics once student rollouts move beyond regions reliably supported by the teacher \cite{liu2026tgpo}. Existing approaches therefore explore uncertainty-aware weighting, entropy-guided optimization, rollout-group coordination, and curriculum-based strategies to stabilize OPD optimization \cite{zheng2026scope,jin2026entropy,yu2026mopd,xu2026paced}.

Despite these advances, existing methods largely coordinate KD and RL through globally fixed schedules or shared optimization objectives, without explicitly modeling when the learning signal should transition from distillation to reinforcement learning. This limitation is especially important in vertical-domain few-shot settings, where OPD plays a different role from conventional refinement-oriented distillation. Since the student initially lacks sufficient domain-specific semantic priors, OPD must first support domain knowledge acquisition before reinforcement learning can effectively refine downstream predictions. As training progresses, however, the marginal benefit of continued OPD gradually diminishes, suggesting that the dominant learning signal should shift from teacher-guided acquisition to reward-driven refinement.

A key challenge is that this KD-to-RL transition point varies substantially across samples. We observe that samples with higher teacher-side perplexity often continue to benefit from prolonged OPD optimization, indicating that they still require teacher-guided acquisition. In contrast, samples with lower teacher-side perplexity typically exhibit limited improvement under continued distillation and can enter reward-driven refinement earlier. These observations suggest that teacher-evaluated sequence perplexity provides a useful signal for estimating sample-level acquisition status. Consequently, KD-to-RL coordination should be governed by sample-level optimization dynamics rather than globally synchronized training stages.

Motivated by these observations, we propose PIVOT (\textbf{P}erplexity-\textbf{I}nformed Transition \textbf{O}ptimization), a dynamic KD-to-RL transition framework that adaptively coordinates OPD-based distillation and GRPO-based refinement \cite{shao2024deepseekmathpushinglimitsmathematical}. PIVOT progressively transitions samples from OPD to GRPO according to teacher-evaluated sequence perplexity: samples with more stable teacher-aligned rollouts enter GRPO refinement earlier, while more difficult samples remain under OPD for continued domain knowledge acquisition. In this way, PIVOT avoids both premature reinforcement learning on under-acquired samples and inefficient over-distillation on samples that are already ready for refinement.

Experiments on Banking77 \cite{casanueva-etal-2020-efficient} and HWU64 \cite{liu2019benchmarkingnaturallanguageunderstanding} show that PIVOT consistently outperforms continued OPD and globally synchronized OPD$\rightarrow$GRPO baselines under the same number of post-warm-up student optimization steps. Further analysis demonstrates that progressive sample-level transition yields more stable optimization dynamics and stronger downstream refinement effectiveness than fixed two-stage transition schedules.

Our contributions are summarized as follows:

\begin{itemize}
    \item We show that prolonged teacher-guided distillation does not necessarily yield better downstream GRPO refinement under vertical-domain few-shot adaptation.

    \item We propose PIVOT, a progressive sample-level KD-to-RL transition framework that adaptively schedules OPD and GRPO according to teacher-evaluated sequence perplexity.

    \item Experiments on Banking77 and HWU64 show that PIVOT consistently outperforms continued OPD and globally synchronized OPD$\rightarrow$GRPO under the same optimization steps while yielding more stable optimization dynamics.
\end{itemize}


\begin{figure*}[!tp]
    \centering
    \includegraphics[width=1\linewidth]{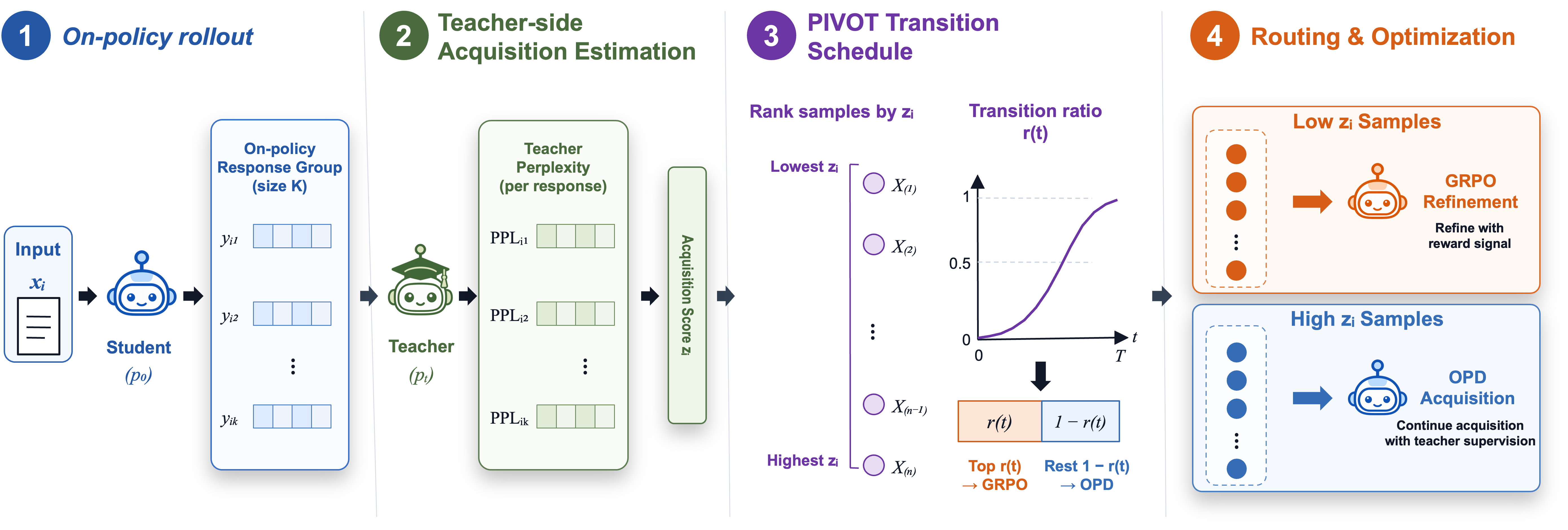
    }  
    \vspace{-7mm} 
    \caption{
    \textbf{Overview of PIVOT.}
    PIVOT estimates sample-level acquisition states using teacher-side perplexity over on-policy student rollouts, and progressively performs asynchronous KD-to-RL transition by routing low-perplexity samples to GRPO refinement while retaining high-perplexity samples under continued OPD acquisition.
    }
    \label{fig:method-overview}
    \vspace{-5mm} 

\end{figure*}

\section{Method}

\subsection{Domain Knowledge Acquisition and Reward-Driven Refinement}

To facilitate domain knowledge acquisition, we first adopt On-Policy Distillation (OPD), which continuously aligns teacher supervision with the student’s evolving rollout distribution.
Specifically, for input $x_i$, the student first generates a response:
\begin{equation}
y_i \sim p_\theta(\cdot|x_i),
\end{equation}
and the student is optimized to match the teacher distribution along on-policy generated trajectories:
\begin{equation}
\small
\mathcal L_{\text{OPD}}
=
\mathbb E_{x_i\sim\mathcal D,\;
y_i\sim p_\theta(\cdot|x_i)}
\left[
\sum_{t=1}^{|y_i|}
D_{\mathrm{KL}}
\!\left(
p_{\mathrm T,t}
\,\|\,
p_{\theta,t}
\right)
\right],
\end{equation}
where
\[
p_{\mathrm T,t}
=
p_{\mathrm T}(\cdot|x_i,y_{i,<t}),
\quad
p_{\theta,t}
=
p_\theta(\cdot|x_i,y_{i,<t}).
\]

After OPD-based acquisition, we further introduce GRPO refinement using a group-relative policy optimization objective.
For each input $x_i$, we sample a group of responses from the old policy
$p_{\theta_{\mathrm{old}}}$:
\[
\{y_{i,j}\}_{j=1}^{G}
\sim
p_{\theta_{\mathrm{old}}}(\cdot|x_i),
\]
where $G$ denotes the group size.
Each response receives a scalar reward $R_{i,j}$, and the corresponding group-normalized advantage is computed as
\[
A_{i,j}
=
\frac{
R_{i,j}-\mu_i
}{
\sigma_i+\delta
},
\]
where $\mu_i$ and $\sigma_i$ denote the mean and standard deviation of rewards within the sampled response group for input $x_i$, respectively, and $\delta$ is a small constant for numerical stability.

For each generated token, the policy ratio is defined as
\[
r_{i,j,t}(\theta)
=
\frac{
p_\theta(y_{i,j,t}\mid x_i,y_{i,j,<t})
}{
p_{\theta_{\mathrm{old}}}(y_{i,j,t}\mid x_i,y_{i,j,<t})
}.
\]
The clipped token-level objective is then given by
\begin{equation}
\small
\begin{aligned}
\ell_{i,j,t}^{\text{GRPO}}
=
\min\Big(
& r_{i,j,t}A_{i,j}, \\
& \mathrm{clip}(r_{i,j,t},1-\epsilon,1+\epsilon)A_{i,j}
\Big).
\end{aligned}
\end{equation}

The GRPO loss is defined as:
\begin{equation}
\small
\begin{aligned}
\mathcal L_{\text{GRPO}}
=
-
\mathbb E_{x_i,\{y_{i,j}\}_{j=1}^{G}}
\Bigg[
\frac{1}{G}
\sum_{j=1}^{G}
\sum_{t=1}^{|y_{i,j}|}
\ell_{i,j,t}^{\text{GRPO}}
\Bigg],
\end{aligned}
\end{equation}
where
\[
x_i\sim\mathcal D,
\quad
\{y_{i,j}\}_{j=1}^{G}
\sim
p_{\theta_{\mathrm{old}}}(\cdot|x_i).
\]

GRPO enables stable relative reward optimization over grouped candidate responses without requiring an additional value model.

In our framework, OPD and GRPO serve complementary optimization roles.
OPD supports domain knowledge acquisition through rollout expansion, whereas GRPO refinement further improves response quality once the marginal acquisition gain from continued distillation gradually diminishes.
This distinction motivates our asynchronous KD-to-RL transition scheduling framework.

\subsection{Heterogeneous Acquisition Dynamics and Acquisition-State Estimation}

Different samples exhibit substantially different acquisition dynamics throughout optimization. Some samples rapidly approach rollout-coverage saturation, whereas others continue benefiting from prolonged OPD acquisition. Consequently, the optimal KD-to-RL transition point varies substantially across samples.

Samples approaching rollout-coverage saturation are better suited for earlier GRPO refinement, while others continue benefiting from OPD-based acquisition. To estimate such acquisition dynamics online, we use teacher-side perplexity over student-generated responses. Lower perplexity indicates that the current student policy has already approached rollout-coverage saturation, whereas higher perplexity suggests that the student policy still benefits from continued acquisition under teacher supervision.

For each input sample $x_i$, the student first performs on-policy rollout and generates a response group:
\begin{equation}
\mathcal Y_i=\{y_{i1},y_{i2},...,y_{iK}\}.
\end{equation}

The teacher model then evaluates each generated response using normalized sequence-level perplexity:
\begin{equation}
\ell_{ij}
=
-\frac{1}{|y_{ij}|}
\sum_{m=1}^{|y_{ij}|}
\log p_{\mathrm T}(y_{ij,m}\mid x_i, y_{ij,<m}),
\end{equation}
\begin{equation}
\mathrm{PPL}_{ij}
=
\exp(\ell_{ij}),
\end{equation}
where $|y_{ij}|$ denotes the response length.

Based on the candidate-level perplexities, we further compute a prompt-level acquisition-state score:
\begin{equation}
u_i
=
\frac{1}{K}
\sum_{j=1}^{K}
\mathrm{PPL}_{ij}.
\end{equation}



A larger $u_i$ indicates that the sample still benefits from continued acquisition under teacher supervision, whereas smaller values suggest that rollout-coverage saturation is gradually approached.
Since absolute perplexity scores are not directly comparable across prompts, we normalize them within each mini-batch:
\begin{equation}
z_i
=
\frac{u_i-\mu_B}{\sigma_B+\epsilon},
\end{equation}
where $\mu_B$ and $\sigma_B$ denote the batch-level mean and standard deviation, respectively.
We use the normalized score $z_i$ as the acquisition state score for routing.

Lower $z_i$ indicates relatively lower teacher-side perplexity within the current mini-batch, suggesting that the sample has acquired sufficient domain knowledge under OPD optimization.
In contrast, higher $z_i$ suggests that the student rollouts remain less aligned with teacher-supported generation patterns and therefore still benefit from continued OPD acquisition.

\subsection{Asynchronous KD-to-RL Transition Scheduling}

Instead of adopting a globally fixed KD-to-RL schedule, PIVOT progressively transitions samples from OPD acquisition to GRPO refinement according to their acquisition state scores.

Specifically, during training, we maintain a transition ratio:
\begin{equation}
r(t)\in[0,1],
\end{equation}
where $t$ denotes the current training step. The transition ratio is progressively increased following a cosine scheduling strategy:
\begin{equation}
r(t)
=
\frac{1}{2}
\left(
1-\cos\frac{\pi t}{T}
\right),
\end{equation}
where $T$ denotes the total number of training steps.

At each training step, samples are ranked according to their normalized acquisition-state scores $z_i$.
Samples with lower acquisition-state scores are considered closer to rollout-coverage saturation and are therefore transitioned earlier into GRPO refinement.
Specifically, we progressively route the lowest-ranked $r(t)$ proportion of samples into GRPO optimization, while the remaining samples continue OPD-based acquisition.
Let $\mathcal S_t$ denote this selected GRPO subset at training step $t$.

Let $\mathcal B_t$ denote the mini-batch at training step $t$.
We define a binary routing indicator $m_i=\mathbf{1}(x_i\in\mathcal S_t)$, where $m_i=1$ indicates that sample $x_i$ is routed to GRPO refinement and $m_i=0$ indicates that it remains under OPD acquisition.
The training objective at step $t$ is then defined as
\begin{equation}
\small
\mathcal L_t
=
\frac{1}{|\mathcal B_t|}
\sum_{x_i\in \mathcal B_t}
\left[
m_i \mathcal L^{(i)}_{\mathrm{GRPO}}
+
(1-m_i)\mathcal L^{(i)}_{\mathrm{OPD}}
\right],
\end{equation}
where $\mathcal L^{(i)}_{\mathrm{GRPO}}$ and $\mathcal L^{(i)}_{\mathrm{OPD}}$ denote the sample-level GRPO and OPD losses for $x_i$, respectively.

In this way, PIVOT enables asynchronous KD-to-RL transition according to acquisition dynamics rather than globally synchronized optimization stages. Samples approaching rollout-coverage saturation transition earlier into GRPO refinement, while other samples continue benefiting from OPD-based acquisition. This progressive transition strategy improves optimization efficiency and downstream decision quality compared with globally fixed KD-to-RL schedules.

\section{Experiments}

\subsection{Experimental Setup}

\paragraph{Datasets and few-shot setting.}
We evaluate our method on Banking77 and HWU64.
For Banking77, we construct a class-balanced 5-shot training set by randomly sampling 5 examples per intent from the original training split, resulting in 385 training instances over 77 intents.
For HWU64, we construct a class-balanced 10-shot training set by sampling 10 examples per intent, resulting in 640 training instances over 64 intents.
We use the official test splits for evaluation.
The few-shot sampling seed is fixed to 42 for all methods.
Both datasets contain fine-grained user intents and are used to evaluate few-shot domain adaptation under limited labeled supervision.

\paragraph{Models.}
Unless otherwise specified, we use Qwen2.5-0.5B-Instruct as the student model and Qwen2.5-7B-Instruct as the teacher model.
We further evaluate whether the proposed method remains effective when scaling the student model from Qwen2.5-0.5B-Instruct to Qwen2.5-1.5B-Instruct.
For stronger-teacher experiments on Banking77, we replace the default teacher with DianJin-32B \cite{zhu2025dianjinr1evaluatingenhancingfinancial}.
The teacher model is kept frozen throughout training and is only used to provide supervision signals for distillation and teacher-side perplexity estimation.

\begin{figure*}[t]
    \centering
    \includegraphics[width=0.9\linewidth]{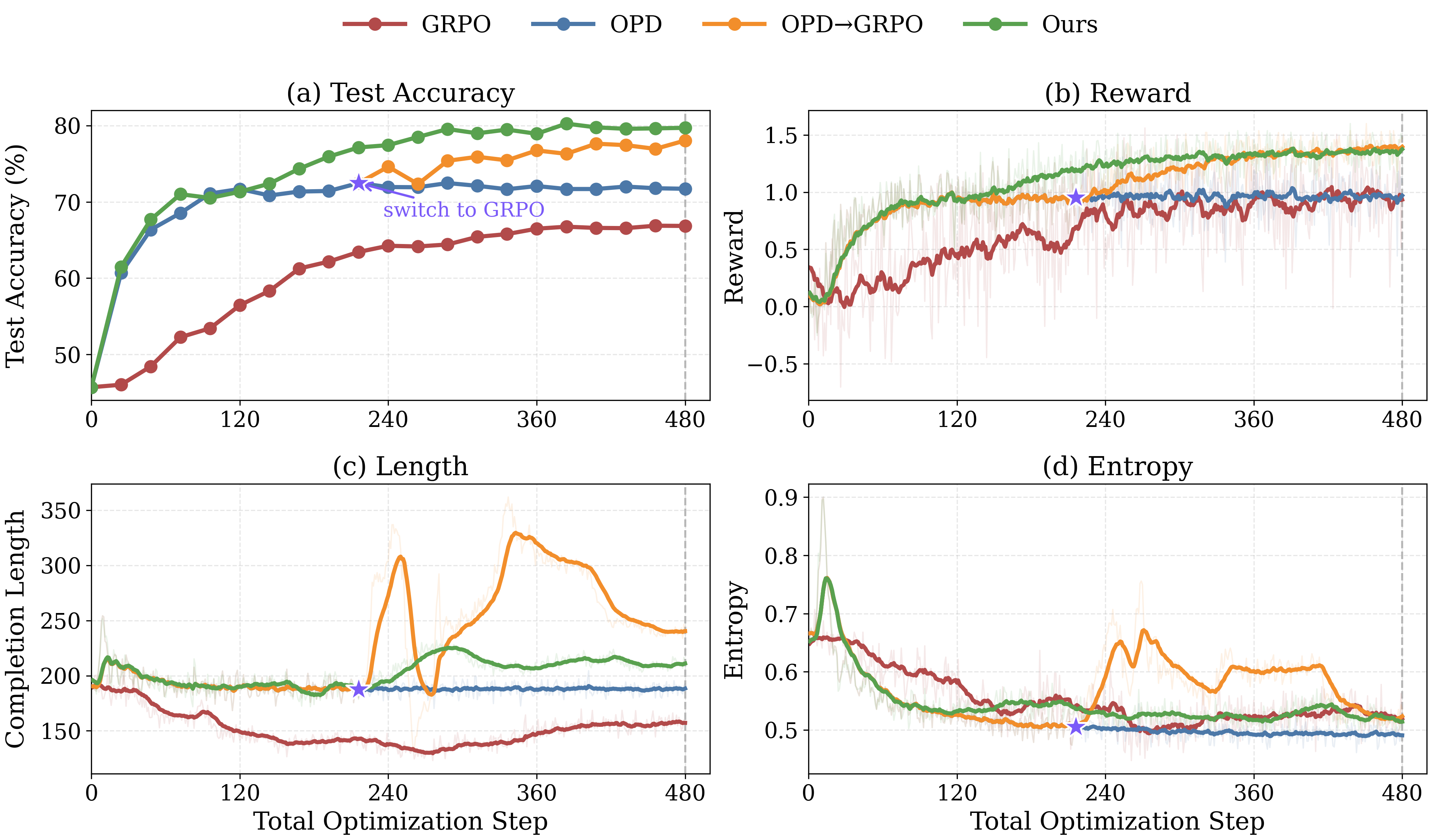}
    \caption{
    Training dynamics on Banking77 under aligned total optimization steps.
    For globally synchronized OPD$\rightarrow$GRPO, the x-axis denotes cumulative optimization steps across the OPD and GRPO stages, and the purple star marks the global transition point.
    Compared with continued OPD and globally synchronized OPD$\rightarrow$GRPO, our progressive sample-level transition achieves higher final accuracy with substantially smoother reward, entropy, and completion-length dynamics.
    }
    \label{fig:training-diagnostics}
\end{figure*}

\vspace{-3mm}

\paragraph{Baselines.}
We compare with the following baselines.
\textbf{Teacher} performs zero-shot prompt-only inference using the teacher model.
\textbf{SeqKD} fine-tunes the student on teacher-generated full responses under the same instruction prompt \cite{kim2016sequence}.
\textbf{SFT$\rightarrow$GRPO} first lightly fine-tunes the student using gold labels and then applies GRPO refinement.
\textbf{DataAug-SFT} fine-tunes the student on a teacher-generated and teacher-filtered augmented training set constructed from the few-shot Banking77 split.
We include this baseline on Banking77, where the augmented set is constructed from the same 5-shot training split.
\textbf{OPD} trains the student with OPD for the full optimization budget.
\textbf{OPD$\rightarrow$GRPO} first performs OPD and then globally switches all samples to GRPO at a fixed transition checkpoint.
    
\paragraph{Training protocol.}
Unless otherwise specified, all trainable student-side methods perform 480 post-warm-up student parameter-update steps.
Following prior OPD practice \cite{li2026unifyinggrouprelativeselfdistillationpolicy}, GRPO, OPD, OPD$\rightarrow$GRPO, and PIVOT methods first share a 2-epoch off-policy SFT warm-up on teacher-generated responses, which is not counted in the 480 steps.
After warm-up, GRPO and OPD use all 480 steps for GRPO refinement and OPD optimization, respectively.
OPD$\rightarrow$GRPO globally switches from OPD to GRPO at the peak OPD checkpoint, while PIVOT progressively routes samples from OPD to GRPO according to teacher-evaluated sequence perplexity.


\begin{table}[t]
\centering
\small
\setlength{\tabcolsep}{6pt}
\begin{tabular}{lcc}
\toprule
\textbf{Method} & \textbf{Banking77} & \textbf{HWU64} \\
\midrule
Teacher             & 73.25 & 79.65 \\
SeqKD               & 58.70 & 70.17 \\
GRPO                & 67.66 & 77.60 \\
OPD       & 72.47 & 79.55 \\
OPD$\rightarrow$GRPO & 78.05 & 83.09 \\
\midrule
\textbf{PIVOT}      & \textbf{80.26} & \textbf{85.04} \\
\bottomrule
\end{tabular}
\caption{
Main results under 480 post-warm-up student optimization steps.
All results report test-set classification accuracy.
}
\label{tab:main_results}
\end{table}

\paragraph{Reward and evaluation.}
For GRPO-based methods, each sampled response is parsed into a predicted intent label from the field ``Intent Category''.
The main reward is based on whether the parsed label matches the gold intent.
We additionally use lightweight auxiliary rewards to encourage valid formatting and consistency between the final answer and the listed candidate labels.
Advantages are normalized within each response group.
At evaluation time, we use greedy decoding and compute micro accuracy on the official test split.
Invalid or unparsable responses are counted as incorrect.

\paragraph{Progressive routing.}
For our method, we compute teacher-evaluated length-normalized sequence perplexity over student-generated responses and average the scores within each response group.
The resulting group-level scores are normalized within each mini-batch using z-score normalization.

After an initial OPD warm-up stage, a cosine schedule progressively increases the proportion of samples transitioned to GRPO throughout training.
At each step, samples with lower normalized perplexity are transitioned earlier to GRPO refinement, while higher-perplexity samples continue OPD optimization.

\subsection{Main Results}

\begin{table*}[t]
\centering
\small
\setlength{\tabcolsep}{6pt}
\begin{tabular}{llccccc}
\toprule
\textbf{Dataset} 
& \textbf{Setting} 
& \textbf{Student} 
& \textbf{Teacher} 
& \textbf{OPD$\rightarrow$GRPO} 
& \textbf{PIVOT} 
& \textbf{$\Delta$} \\
\midrule
Banking77 
& Default 
& 0.5B 
& Qwen2.5-7B 
& 78.05 
& \textbf{80.26} 
& +2.21 \\

Banking77 
& Larger student 
& 1.5B 
& Qwen2.5-7B 
& 80.00 
& \textbf{82.82} 
& +2.82 \\

Banking77 
& Stronger teacher 
& 0.5B 
& DianJin-32B 
& 80.30 
& \textbf{81.33} 
& +1.03 \\
\midrule
HWU64 
& Default 
& 0.5B 
& Qwen2.5-7B 
& 83.09 
& \textbf{85.04} 
& +1.95 \\

HWU64 
& Larger student 
& 1.5B 
& Qwen2.5-7B 
& 82.53 
& \textbf{85.97} 
& +3.44 \\

HWU64 
& Stronger teacher 
& 0.5B 
& Qwen3-32B 
& 84.39 
& \textbf{86.71} 
& +2.32 \\
\bottomrule
\end{tabular}
\caption{
Scaling analysis across different student and teacher model configurations.
$\Delta$ denotes the improvement of PIVOT over OPD$\rightarrow$GRPO.
}
\vspace{-3mm}
\label{tab:scaling}
\end{table*}

Table~\ref{tab:main_results} reports the main results under the same number of post-warm-up student optimization steps.
Compared with continued OPD, globally synchronized OPD$\rightarrow$GRPO already yields substantial improvement on both datasets, suggesting that downstream GRPO refinement is important beyond prolonged imitation alone.
However, PIVOT further improves over OPD$\rightarrow$GRPO across both Banking77 and HWU64, indicating that globally fixed transition schedules remain suboptimal for coordinating OPD and GRPO during training.

Interestingly, PIVOT also surpasses teacher inference on both datasets, suggesting that progressive sample-level transition can produce a stronger task-specialized student policy under few-shot adaptation.
We additionally compare against a teacher-generated data augmentation baseline on Banking77 in Appendix~\ref{sec:data-augmentation}.
Although DataAug-SFT improves over standard SFT, it remains below both continued OPD and PIVOT, suggesting that the gains are not solely attributable to additional teacher-generated training data.

\subsection{Optimization Dynamics}

Figure~\ref{fig:training-diagnostics} compares test accuracy, reward, entropy, and completion length under aligned total optimization steps.
Our method consistently achieves the highest final test accuracy and surpasses globally synchronized OPD$\rightarrow$GRPO before the global transition point, suggesting that progressive sample-level transition improves downstream refinement effectiveness under the step-matched protocol.

Although globally synchronized OPD$\rightarrow$GRPO eventually reaches comparable reward levels, it exhibits noticeably unstable optimization behavior near the transition stage, including abrupt completion-length increase and entropy spikes.
In contrast, our method maintains substantially smoother entropy and completion-length dynamics throughout training, indicating that progressively transferring low-perplexity samples into GRPO avoids the optimization discontinuity introduced by globally synchronized switching.
Interestingly, our method achieves better final accuracy despite similar late-stage reward, suggesting that progressive transition improves the effectiveness of reward-driven refinement rather than merely increasing reward magnitude.

\subsection{Scaling Analysis}

We further evaluate PIVOT across different student and teacher model configurations.
As shown in Table~\ref{tab:scaling}, PIVOT consistently improves over OPD$\rightarrow$GRPO across all evaluated settings, including larger student models and stronger teacher models. Notably, the improvement remains consistent even when scaling the student from 0.5B to 1.5B parameters, suggesting that the benefit of progressive sample-level transition is not tied to a specific student capacity.
Similarly, PIVOT continues to outperform OPD$\rightarrow$GRPO under stronger teacher models, indicating that adaptive transition remains beneficial even when teacher supervision quality improves substantially.

\subsection{Ablation and Routing Analysis}



\begin{table}[t]
\centering
\small
\setlength{\tabcolsep}{4pt}
\begin{tabular}{lccc}
\toprule
\textbf{Dataset} 
& \textbf{Low-PPL} 
& \textbf{Random} 
& \textbf{High-PPL} \\
\midrule
Banking77 
& \textbf{80.26} 
& 78.31 
& 77.27 \\

HWU64 
& \textbf{85.04} 
& 84.11 
& 83.46 \\
\bottomrule
\end{tabular}
\caption{
Routing-rule ablation under progressive OPD-to-GRPO transition.
Column names indicate which samples are preferentially routed to GRPO earlier during training.
}
\label{tab:routing_rule}
\end{table}

\begin{table}[t]
\centering
\small
\setlength{\tabcolsep}{10pt}
\begin{tabular}{lc}
\toprule
\textbf{Transition Strategy} & \textbf{ACC} \\
\midrule
Global OPD$\rightarrow$GRPO & 78.05 \\
Linear progressive routing & 79.58 \\
Quadratic progressive routing & 79.19 \\
\textbf{Cosine progressive routing} & \textbf{80.26} \\
\bottomrule
\end{tabular}
\caption{
Transition-schedule ablation on Banking77 under the same number of student optimization steps.
}
\vspace{-4mm} 

\label{tab:schedule_ablation}
\end{table}

\begin{figure}[t]
\centering
\includegraphics[width=\linewidth]{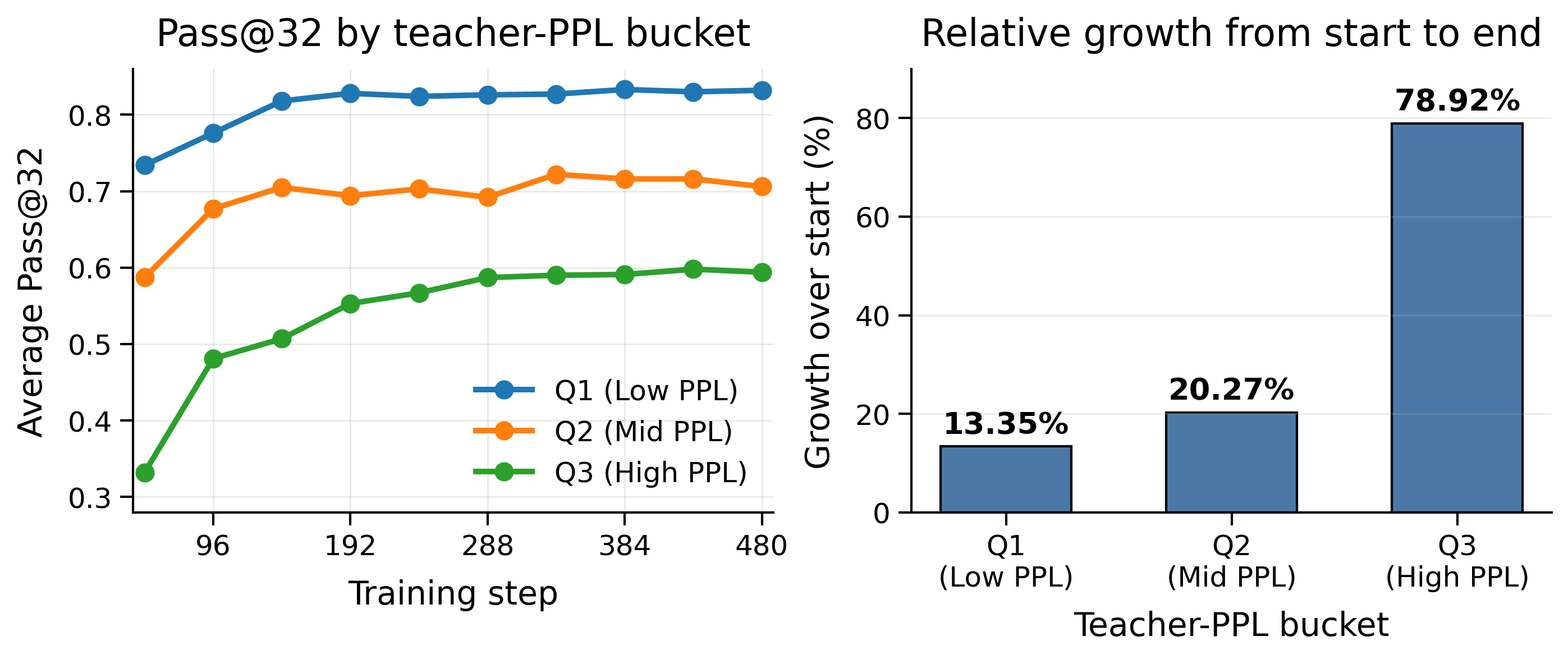}

\vspace{-2mm} 

\caption{
Pass@32 performance grouped by teacher-side perplexity buckets during OPD training.
High-perplexity samples exhibit substantially larger relative improvement throughout training, whereas low-perplexity samples saturate much earlier, suggesting that the optimal transition timing from OPD to RL varies across samples.
}
\label{fig:ppl-analysis}
\end{figure}

\vspace{-2mm}

We further analyze the design choices of progressive OPD-to-GRPO transition, including both the transition schedule and the routing strategy. As shown in Table~\ref{tab:schedule_ablation}, all progressive routing variants consistently outperform globally synchronized OPD$\rightarrow$GRPO, indicating that asynchronous sample-level transition is more effective than switching all samples at a fixed checkpoint. Among different schedules, cosine routing achieves the strongest overall performance, suggesting that gradually increasing the GRPO transition ratio throughout training provides a better balance between continued OPD optimization and downstream refinement. Table~\ref{tab:routing_rule} further shows that the best results are achieved by retaining high-PPL samples in OPD while transitioning low-PPL samples to GRPO earlier, suggesting that different samples benefit from substantially different transition timing during training. Figure~\ref{fig:ppl-analysis} further supports this behavior: lower-PPL samples rapidly approach rollout-quality saturation, whereas higher-PPL samples continue improving under prolonged OPD optimization. These observations support teacher-side perplexity as an effective signal for progressive sample-level KD-to-RL transition.

\section{Discussion}

\subsection{Distillation Quality and Refinement Readiness Are Not Necessarily Aligned}

Our results suggest that stronger distillation quality does not necessarily imply better downstream refinement readiness.
Existing two-stage adaptation pipelines often assume that continued teacher alignment monotonically improves the initialization quality for subsequent preference optimization.
However, our experiments show that the strongest downstream GRPO performance frequently emerges before full OPD convergence.

One possible explanation is that prolonged imitation gradually narrows the student rollout distribution around teacher-preferred responses.
Although this improves teacher alignment, it may also reduce the diversity and observability of informative candidate responses required for downstream GRPO refinement.
As a result, stronger imitation quality does not always translate into more effective downstream optimization.

These observations suggest that distillation and downstream refinement correspond to partially different optimization objectives, and that maximizing teacher alignment alone may not be sufficient for effective refinement-oriented adaptation.

\subsection{Implications for Adaptive Distillation-to-Refinement Transition}

Our findings further suggest that the transition from distillation to downstream refinement should not be globally synchronized across all samples.
Different samples exhibit substantially different optimization dynamics during OPD training, implying that the optimal transition timing may vary significantly across samples.

In particular, low teacher-side perplexity consistently correlates with more stable rollout behavior and stronger pass@32 performance throughout training, suggesting that these samples are more suitable for earlier refinement.
In contrast, high-perplexity samples often continue benefiting from prolonged OPD optimization before downstream GRPO becomes effective.

Although our study focuses on OPD and GRPO under vertical-domain few-shot classification, similar interactions may arise more broadly in distillation-based adaptation settings involving iterative imitation and downstream preference optimization.
More generally, our results highlight the importance of considering not only teacher alignment quality, but also how optimization dynamics influence downstream refinement behavior throughout training.

\section{Related Work}

\subsection{On-Policy Distillation and Adaptive Optimization}

Knowledge distillation has become a widely adopted paradigm for transferring capabilities from large language models to smaller models under constrained supervision or computational settings \cite{hinton2015distilling,sanh2019distilbert}. Recent work has increasingly explored On-Policy Distillation (OPD), where teacher supervision is continuously aligned with the student’s evolving rollout distribution during optimization, improving exposure alignment during autoregressive generation \cite{agarwal2024onpolicy,gu2024minillm}. 

Recent OPD studies further explore adaptive optimization strategies for stabilizing supervision under heterogeneous rollout conditions. Existing approaches include entropy-aware distillation objectives \cite{jin2026entropy}, uncertainty-calibrated supervision weighting \cite{zheng2026scope}, rollout-group coordination \cite{yu2026mopd}, and competence-based optimization schedules \cite{xu2026paced}. Recent analyses of OPD failure modes further show that teacher guidance can become unreliable on student-generated prefixes, especially when rollouts move away from prefixes commonly visited by the teacher \cite{fu2026revisitingonpolicydistillationempirical}. 

These studies suggest that the effectiveness of teacher supervision can vary substantially throughout OPD optimization under heterogeneous rollout conditions. Our work is related to these adaptive OPD optimization methods but focuses on a different problem setting. Instead of improving supervision quality within OPD itself, we study how acquisition dynamics should govern the transition between distillation and downstream RL refinement during vertical-domain few-shot adaptation.

\subsection{KD-RL Coordination and Adaptive Transition}

Preference- and reward-based optimization methods such as RLHF, DPO, and GRPO have become important paradigms for improving downstream behavior beyond supervised imitation \cite{ouyang2022training,rafailov2023direct,shao2024deepseekmath}. 
In particular, GRPO improves optimization stability through group-relative reward normalization and has shown strong effectiveness in reasoning-oriented refinement settings \cite{shao2024deepseekmath}. 
Recent work further applies GRPO to intent detection by combining it with chain-of-thought reasoning and reward-based curriculum sampling, showing that reward-driven optimization can improve intent understanding and generalization beyond supervised fine-tuning \cite{feng2025improvinggeneralizationintentdetection}.

Motivated by the complementary roles of distillation and downstream refinement, recent work has increasingly explored how knowledge distillation and reinforcement learning should be coordinated throughout optimization. RLKD \cite{xu2025rlkd} and KDRL \cite{xu2025kdrl} show that reinforcement learning can complement teacher-guided distillation by improving exploration and downstream refinement beyond imitation alone. SuperRL \cite{liu2025superrl} further incorporates supervised signals into reinforcement learning to improve reasoning optimization. G-OPD \cite{yang2026gopd} further establishes a theoretical connection between OPD and KL-regularized reinforcement learning, suggesting that OPD itself can be interpreted as a dense reward optimization process. CoDistill-GRPO \cite{kwon2026codistill} and related hybrid optimization frameworks \cite{li2026unifying} additionally demonstrate that jointly coordinating distillation and reinforcement learning can improve optimization efficiency and rollout quality during training.

However, existing approaches primarily rely on globally shared objectives or fixed optimization schedules, without explicitly modeling sample-level transition dynamics between distillation and reinforcement learning.

\section{Conclusion}

In this paper, we propose PIVOT, a progressive sample-level KD-to-RL transition framework for vertical-domain few-shot adaptation. PIVOT adaptively schedules OPD and GRPO according to teacher-evaluated sequence perplexity rather than globally fixed transition schedules. Experiments on Banking77 and HWU64 show that PIVOT consistently outperforms continued OPD and globally synchronized OPD$\rightarrow$GRPO under the same number of post-warm-up student optimization steps, while yielding more stable optimization dynamics during training. Our results suggest that adaptive sample-level transition provides an effective alternative to fixed two-stage distillation and refinement schedules.

\clearpage

\section{Limitations}

Our study has several limitations. First, PIVOT relies on teacher-evaluated sequence perplexity as a heuristic routing signal, and we do not provide a theoretical characterization of why perplexity best reflects refinement readiness across different optimization settings. Second, our experiments focus on vertical-domain few-shot classification with relatively structured outputs and reward signals. Whether similar transition dynamics extend to more complex generation tasks remains an open question. Third, our comparisons align the number of student optimization steps rather than total training compute. Because PIVOT continues teacher-side perplexity evaluation for samples routed to GRPO, it introduces additional teacher inference overhead relative to the globally synchronized baseline after its transition. The reported results should therefore be interpreted as step-matched rather than strictly compute-matched comparisons. We leave broader theoretical analysis, stricter compute-matched evaluation, and evaluation on more diverse tasks for future work.

\section*{AI Assistance Statement}

Generative AI tools, including ChatGPT and Gemini, were used to assist with language editing during the preparation of this work. All generated content was reviewed and verified by the authors. The authors take full responsibility for the accuracy, integrity, and originality of the final manuscript.

\bibliography{custom}

\clearpage

\clearpage
\appendix

\section{Implementation Details}
\label{app:implementation}

\subsection{Dataset Construction}

We construct class-balanced few-shot training sets from the original training splits.
Banking77 uses 5 examples per intent, resulting in 385 training examples over 77 intents, while HWU64 uses 10 examples per intent, resulting in 640 training examples over 64 intents.
The sampling seed is fixed to 42.
We use the official test splits for evaluation and report micro accuracy.
Unless otherwise stated, all experiments are conducted on the same sampled few-shot split.
Following prior few-shot adaptation settings, the main results are reported on a single split because OPD and GRPO training are computationally expensive.

\subsection{Prompt Templates}

All methods use task-specific instruction prompts containing the full label set.
Banking77 additionally includes natural-language intent definitions, while HWU64 uses only the intent label list.
SFT-ZeroShot uses a simplified label-only supervision format, while teacher inference, SeqKD, OPD, GRPO, and PIVOT use the structured reasoning prompts shown in Tables~\ref{tab:banking_prompt} and~\ref{tab:hwu_prompt}.

\begin{table*}[t]
\centering
\caption{Prompt template for Banking77.}
\label{tab:banking_prompt}
\begin{minipage}{0.96\textwidth}
\begin{lstlisting}[style=promptstyle]
You are an expert intent classifier for a banking customer service bot.
Your task is to classify the user's query into exactly one of the following 77 intent categories.

### Intent Definitions
{definitions_str}

### Instructions
1. **Analyze Key Elements**: Identify the core action (e.g., "top up", "transfer") and the object.
2. **List Candidates (Crucial)**:
 - Before deciding, list 3-5 category names from the definitions that are most relevant to the query.
3. **Compare & Filter**:
 - Compare the candidates against the specific details of the query.
 - Explicitly rule out the ones that don't fit perfectly.
4. **Final Conclusion**: Select the single best matching category.

### Response Format
You must strictly follow this format:
**Analysis:** [Brief analysis of key elements]
**Candidates:** [List 3-5 potential category names here]
**Reasoning:** [Your comparison and filtering process]
**Intent Category:** [The exact category name]
\end{lstlisting}
\end{minipage}
\vspace{1mm}
\caption*{\small \texttt{\{definitions\_str\}} contains the full set of Banking77 intent labels and their natural-language definitions.}
\end{table*}

\begin{table*}[t]
\centering
\caption{Prompt template for HWU64.}
\label{tab:hwu_prompt}
\begin{minipage}{0.96\textwidth}
\begin{lstlisting}[style=promptstyle]
You are an expert intent classifier for a multi-domain virtual assistant.
Your task is to classify the user's query into exactly one of the following {n_classes} intent categories.

### Intent Categories
{labels_str}

### Instructions
1. **Analyze Key Elements**: Identify the core action and object in the query.
2. **List Candidates (Crucial)**:
 - Before deciding, list 3-5 category names from the intent categories that are most relevant to the query.
3. **Compare & Filter**:
 - Compare the candidates against the specific details of the query.
 - Explicitly rule out the ones that don't fit perfectly.
4. **Final Conclusion**: Select the single best matching category.

### Response Format
You must strictly follow this format:
**Analysis:** [Brief analysis of key elements]
**Candidates:** [List 3-5 potential category names here]
**Reasoning:** [Your comparison and filtering process]
**Intent Category:** [The exact category name]
\end{lstlisting}
\end{minipage}
\vspace{1mm}
\caption*{\small \texttt{\{labels\_str\}} contains the full HWU64 intent label set.}
\end{table*}

Here, \texttt{\{labels\_str\}} contains the full HWU64 intent label set.

\subsection{Output Parsing}

We parse the final prediction from the field ``Intent Category''.
The first valid label appearing after ``Intent Category:'' is matched against the task-specific label set.
If multiple valid labels appear after this field, the first valid match is used.
If no valid label can be matched, the prediction is counted as incorrect.
No manual correction is applied.

\subsection{Baseline Implementations}

GRPO and OPD-based methods use the same 2-epoch off-policy SFT warm-up on teacher-generated responses before on-policy optimization.

\paragraph{Teacher.}
The frozen teacher model performs zero-shot prompt-only inference using the same task prompt as evaluation.

\paragraph{SFT-ZeroShot.}
The student is fine-tuned on the few-shot gold training set using label-only supervision targets.

\paragraph{SeqKD.}
For each few-shot training example, the teacher generates one full response under the same instruction prompt.
The student is then fine-tuned on the resulting teacher-generated responses.

\paragraph{GRPO.}
The student is optimized directly with GRPO after the off-policy warm-up stage.

\paragraph{DataAug-SFT.}
The student is fine-tuned on a teacher-generated augmented training set with teacher-based filtering constructed from the same Banking77 5-shot split.
This baseline is included only on Banking77.

\paragraph{OPD.}
The student is trained with OPD for all 480 post-warm-up student optimization steps.

\paragraph{OPD$\rightarrow$GRPO.}
The student is first trained with OPD and then globally switched to GRPO at a fixed transition checkpoint.
The best transition checkpoint is selected within the same total of 480 post-warm-up student optimization steps.

\subsection{Data Augmentation Baseline}
\label{sec:data-augmentation}

We additionally evaluate a teacher-generated data augmentation baseline on Banking77 to test whether the gains of PIVOT can be explained by additional synthetic training data alone.

Starting from the same 5-shot Banking77 split, we construct an augmented training set using intent clustering, teacher-based query generation, and teacher self-filtering.
Semantically similar intents are first grouped into clusters.
For each cluster, the teacher generates additional user queries conditioned on the target intent, intent definition, and several anchor examples.
The generation prompt encourages diverse and label-discriminative queries.
Generated samples are then reclassified by the teacher, and only samples whose predicted labels match the target intent are retained.

Using this pipeline, we generate 5,650 candidate augmented examples, of which 5,119 are retained after teacher self-filtering.
Fine-tuning on the resulting augmented set achieves 71.75 test accuracy on Banking77.
While synthetic data augmentation substantially improves over standard SFT, it remains below both continued OPD and OPD$\rightarrow$GRPO training.
Notably, introducing the OPD-to-GRPO transition yields a large additional gain, suggesting that the effectiveness of PIVOT cannot be explained solely by additional teacher-generated supervision or extended optimization.

\begin{table}[t]
\centering
\small
\setlength{\tabcolsep}{8pt}
\begin{tabular}{lc}
\toprule
\textbf{Method} & \textbf{Banking77} \\
\midrule
SFT-ZeroShot         & 60.30 \\
DataAug-SFT          & 71.75 \\
Continued OPD        & 72.47 \\
OPD$\rightarrow$GRPO & 78.05 \\
PIVOT                & \textbf{80.26} \\
\bottomrule
\end{tabular}
\caption{
Comparison between teacher-generated data augmentation and different OPD/GRPO training strategies on Banking77.
DataAug-SFT is trained on 5,119 teacher-filtered synthetic examples generated from the same 5-shot split.
}
\label{tab:data-augmentation}
\end{table}

\begin{table*}[t]
\centering
\caption{Prompt template for teacher-based data augmentation.}
\label{tab:data-augmentation-prompt}
\begin{minipage}{0.96\textwidth}
\begin{lstlisting}[style=promptstyle]
You are an expert at generating realistic bank customer inquiry data.
Your task is to simulate real banking customers, generating highly conversational
and natural customer service queries that accurately map to specific intents.

I need to augment training data for a highly sensitive bank intent recognition model.
Below is a group of {num_intents} intents that are semantically similar and easily confused.
They share overlapping vocabulary but have critical semantic differences.

For each target intent, you are given:
- the intent name;
- its definition;
- several reference anchor queries.

TASK:
Generate exactly {n_per_intent} completely new customer queries for EACH intent.
Total output = {n_per_intent * num_intents} items.

Strict Constraints:
1. Boundary Exclusivity:
   Each query must be unambiguous and must only map to its declared intent,
   not to any other intent in this group.

2. Extreme Diversity:
   Vary sentence length, structure, formality, and scenario.
   Use informal speech, colloquialisms, and specific real-life situations.
   Do not simply paraphrase the seed examples.

3. Scenario-driven:
   Imagine concrete real-life situations, such as standing at a supermarket checkout,
   receiving a bank notification, or finding an unexpected charge.

4. No repetition:
   Do not repeat any reference anchor sentence.

Output ONLY a JSON array:
[
  {"intent": "{intent_1}", "text": "..."},
  {"intent": "{intent_2}", "text": "..."}
]
\end{lstlisting}
\end{minipage}
\end{table*}

\subsection{Training Hyperparameters}

Table~\ref{tab:hyperparameters} summarizes the main training hyperparameters used across all experiments. All experiments were conducted on a server with four NVIDIA A100 GPUs.

\begin{table}[t]
\centering
\small
\begin{tabular}{lc}
\toprule
\textbf{Hyperparameter} & \textbf{Value} \\
\midrule
Total optimization steps & 480 \\
Batch size & 256 \\
Rollouts / group size & 8 \\
Warmup ratio & 0.05 \\
Max sequence length & 4096 \\
Max new tokens & 512 \\
Temperature & 1.0 \\
Top-$p$ & 0.9 \\
Top-$k$ & 50 \\
Precision & bf16 \\
Fine-tuning type & full \\
\midrule
Learning rate (OPD) & $1\times10^{-5}$ \\
Learning rate (GRPO / Ours) & $1\times10^{-6}$ \\
KL coefficient (GRPO / Ours) & 0.001 \\
\bottomrule
\end{tabular}
\caption{Training hyperparameters used in all experiments.}
\label{tab:hyperparameters}
\end{table}


\subsection{Reward Function}

For GRPO-based methods, the final intent label is parsed from each generated response.
The reward consists of three components: a correctness reward based on whether the parsed label matches the gold intent (+1 / -1), a candidate-alignment reward encouraging consistency between the final prediction and the listed candidate labels (+0.5 / -0.5), and a lightweight format reward encouraging valid structured outputs (+0.1).
We do not apply an explicit length penalty.
A KL penalty with coefficient 0.001 is used during GRPO optimization, and advantages are normalized within each response group.


\end{document}